\documentclass[journal]{IEEEtran}

\usepackage{xcolor,soul,framed} 

\colorlet{shadecolor}{yellow}
\usepackage[pdftex]{graphicx}
\graphicspath{{../pdf/}{../jpeg/}}
\DeclareGraphicsExtensions{.pdf,.jpeg,.png}

\usepackage[cmex10]{amsmath}
\usepackage{array}
\usepackage{mdwmath}
\usepackage{mdwtab}
\usepackage{eqparbox}
\usepackage{url}
\usepackage{booktabs} 
\newcommand{\eat}[1]{}
\usepackage{bbold}
\usepackage{subfig}
\usepackage{booktabs}
\usepackage{multirow}
\usepackage{siunitx}
\begin{document}
\bstctlcite{IEEEexample:BSTcontrol}
    \title{Snap and Find: Deep Discrete Cross-domain Garment Image Retrieval}
  \author{Yadan Luo, Ziwei Wang, Zi Huang, Yang Yang, Huimin Lu

  \thanks{}
  \thanks{Y. Luo, Z. Wang, Z. Huang are with the School of Information Technology and Electrical Engineering, The University of Queensland, Brisbane, QLD, 4072, Australia (e-mail: lyadanluol@gmail.com; ziwei.wang@uq.edu.au; zhuang@itee.uq.edu.au;).}
  \thanks{Y. Yang is with the School of Computer Science and Engineering, University of Electronic Science and Technology of China, Chengdu 611731, China (e-mail: dlyyang@gmail.com).}%
  \thanks{H. Lu is with the Department of Mechanical and Control Engineering, Kyushu Institute of Technology, Japan (e-mail: dr.huimin.lu@ieee.org).}
}

\markboth{IEEE TRANSACTIONS ON IMAGE PROCESSING, VOL.~60, NO.~12, DECEMBER~2012
}{Luo \MakeLowercase{\textit{et al.}}: Snap and Find: Deep Discrete Garment Image Retrieval}

\maketitle

\begin{abstract}
With the increasing number of online stores, there is a pressing need for intelligent search systems to understand the item photos snapped by customers and search against large-scale product databases to find their desired items. However, it is challenging for conventional retrieval systems to match up the item photos captured by customers and the ones officially released by stores, especially for garment images. To bridge the customer- and store- provided garment photos, existing studies have been widely exploiting the clothing attributes (\textit{e.g.,} black) and landmarks (\textit{e.g.,} collar) to learn a common embedding space for garment representations. Unfortunately they omit the sequential correlation of attributes and consume large quantity of human labors to label the landmarks. In this paper, we propose a deep multi-task cross-domain hashing termed \textit{DMCH}, in which cross-domain embedding and sequential attribute learning are modeled simultaneously. Sequential attribute learning not only provides the semantic guidance for embedding, but also generates rich attention on discriminative local details (\textit{e.g.,} black buttons) of clothing items without requiring extra landmark labels. This leads to promising performance and 306$\times$ boost on efficiency when compared with the state-of-the-art models, which is demonstrated through rigorous experiments on two public fashion datasets. 
\end{abstract}


%
\IEEEpeerreviewmaketitle


\section{Introduction}

Throughout the world people are gradually tired of shopping in their local stores with the problems of parking, long lines, and wobbly shopping carts. Instead, online buying has grown exponentially, characterized by strong consumer demands and a cumulative number and type of goods available. Most of online retail shops feature the ``garment image search'' function, which allows users to submit a photo for looking for its corresponding products.

\eat{
	\begin{figure}[htb!]
		\centering
		\includegraphics[width=0.5\textwidth]{images/figure1.pdf}
		\caption{The proposed framework of cross-domain garment image retrieval. In the embedding space, the positive user-shop image pair is projected closer, while the negative one stays distant.}
		\label{fig:example}
	\end{figure}
}

Product image search, a practical example of cross-domain retrieval, requires to measure the similarity of images that are from two heterogeneous domains, termed the \emph{user domain} and the \emph{shop domain}. The user domain consists of the query images taken by users (\textit{e.g.,} street snap, Instagram post), and the shop domain includes the database images taken by professional photographers.

Cross-domain garment image retrieval is essentially challenging due to (1) large intra-class variance and (2) minor inter-class variance. For the same product, large intra-class variance includes the different lighting conditions, noisy backgrounds, poses, even occlusion and deformation between the user- and shop- image, causing the garment images from different domains are hard to be matched up. Minor inter-class variance is an intrinsic property of garment images. For example, two dresses from different categories can be very similar in color and design yet have a minor difference in the collar's shape, where one is V-shaped and the other one is U-shaped. Given a user garment image with a V-shaped collar, returning the dress with a U-shaped collar is not considered as a correct search result in our scenario. The accuracy of fine-grained garment image search is highly affected by the low discriminations among alike garments, so that it is inevitably tougher than the conventional Content-Based Image Retrieval~(CBIR), where the subtle and local differences are expected to be identified.

\begin{figure*}[htb!]
	\centering
	\includegraphics[width=1\textwidth]{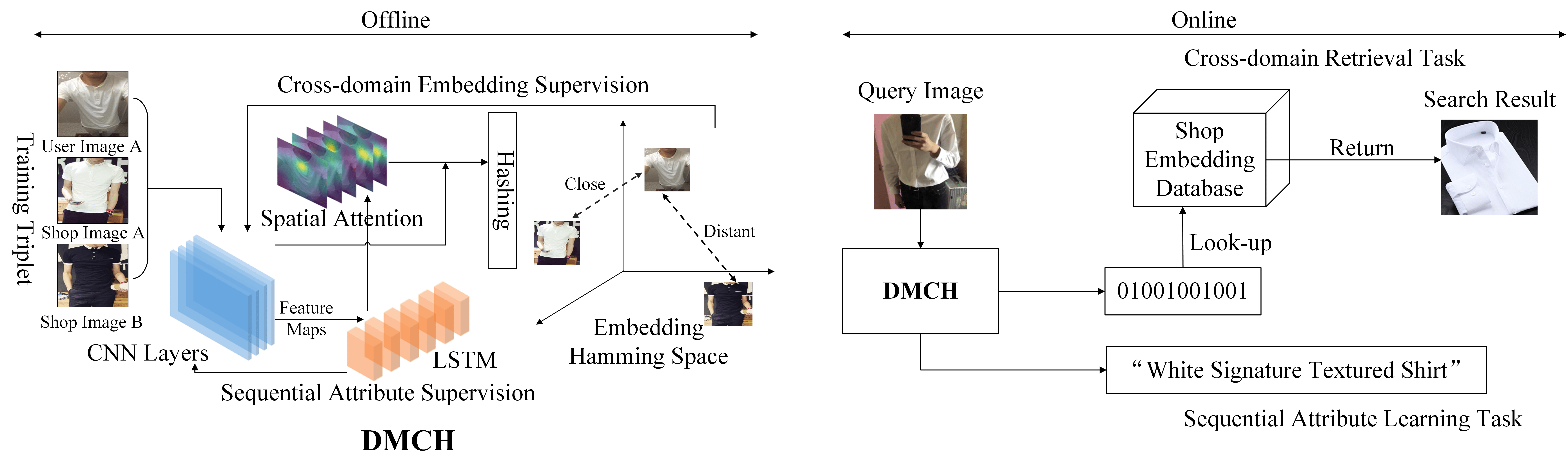}
	\caption{An overview of the proposed framework. During the offline stage, the triplet label (user image A, shop image A and shop image B) and the corresponding sequential attributes guide the DMCH model with spatial attention to generate the binary embedding. At the online stage, given a query user image, the DMCH generate its predicted sequential attributes and its binary code with which the matched shop image can be retrieved from the shop image embedding database.}
	\label{fig:framework}
\end{figure*}

In order to alleviate the above-mentioned problems, emerging literatures have proposed solutions in mainly two views: 1) visual objectness learning; 2) semantic attribute learning. Technically, the objectness learning based methods detect the foreground object and/or its semantic parts~\cite{Deepfashion, shoes} before extracting features, mainly for suppressing discrepancy of the background and enhancing recognition of related local details. Apparently, this method highly relies on the annotated bounding boxes or clothing landmarks for supervised detection, which is time- and labor-consuming. An alternative for detection is to extract saliency or visual attention~\cite{Attention, concept} from raw images. Different from objectness learning, which facilitates fine-grained recognition by detecting visual parts, attribute learning~\cite{DARN, DPH} constructs a latent space between fine-grained labels and low-level features. Attributes of clothing items such as color, texture, fabric or even style assist the model to find the inter-class and intra-class correlation between garment categories. The existing attribute learning models generally address attribute prediction as a multi-label classification problem, where each attribute is considered as a category or class. Actually, each garment image for model training is associated with a sequence of attributes such as ``silk pocket shirt''. However, the sequential information is omitted by the conventional attribute learning models.

In this paper, we propose an end-to-end deep multi-task cross-domain hashing (DMCH) to jointly model the sequential correlation among the clothing attributes and learn the attention-aware visual features of garment images for both cross-domain retrieval and sequential attribute learning tasks. From the linguistic perspective\footnote{https://dictionary.cambridge.org/grammar/british-grammar/about-adjectives-and-adverbs/adjectives-order}, when describing a noun, the adjectives before it are usually in a particular order. For instance, people usually get used to say a ``little'' ``black'' ``dress'' rather than a ``black'' ``little'' ``dress''. Two popular fashion datasets~\cite{DARN, Deepfashion} apply alike order to create the attribute lists, based on which each garment image is described by a sequence of attributes in a certain order. Instead of assigning discrete attributes to a query image, our model considers the sequential correlation among attributes and predicts a sequence of attributes with a meaningful order for each query, where the confidence of each attribute to be selected is largely enhanced. Additionally, most of the existing work plainly aggregate features from different convolutional layers and apply the Euclidean distance measure on the concatenated feature vectors, unavoidably resulting in inefficient query processing and unnecessary storage waste. Learning to hash component in the proposed model successfully addresses these potential issues. Figure 1 gives an overview of the proposed framework, which includes the offline training of the DMCH model and the online process for cross-domain retrieval and attribute generation. The main contributions are summarized below:
\begin{itemize}
	\item A deep discrete embedding framework is proposed to learn discriminative representations of garment images, being supervised by two objectives simultaneously. Cross-domain correlation and visual-semantic association are utilized jointly in garment image representation learning to facilitate both tasks. 
	
	\item The spatial attention of garment images is activated by attribute descriptions, which significantly enhances the recognition on the subtle and local details.
	\item Different from the conventional attribute learning, which is usually positioned as a multi-label classification problem, we open a new direction of sequential attribute learning, where the strong linguistic hint is leveraged.
	
	\item The learned binary embedding of garment images is efficient and effective with strong discriminative power, which enables our algorithm to accelerate 306 $\times$ on querying without compromising the accuracy.
\end{itemize}

\section{Related Work} 

\subsection{Cross-domain Image Retrieval}
Product image search is one type of Content-Based Image Retrieval~(CBIR), which has been widely studied for decades~\cite{shensurvey, bokun}. Hashing based methods~\cite{ITQ,SDH,KSH,RSDH,Zerohash} serve as effective solutions, which transform homogeneous high dimensional samples into similar compact binary codes with which data similarity can be measured by bit-wise xor operation. Recently, end-to-end deep hashing~\cite{DPSH,DBLP:conf/cvpr/Liu0SC17,DBLP:conf/mm/ShenGLYS17} has shown its superiority in both feature representation and embedding quality in comparison with previous two-pipeline supervised approaches.

However, along with the explosive growth of web data, even deep image retrieval is still facing enormous cross-domain challenges, \textit{i.e.,} heterogeneous query-result pairs and fine-grained classes. For instance, using a street snap of a lady in white coat as a query to find the shop image of the exact same item could hardly cut the mustard, especially with the different backgrounds and poses. 

WTBI~\cite{Wtbi} handles the fine-grained classes problem with an easy-to-hard strategy, that means, training a generic network with five main classes followed by subtle fine-tuning for each sub-class individually. DARN~\cite{DARN} has two NIN subnets, one for street domain and one for shop domain, that enable the model to learn the domain-specific representation. To discover the common semantic features, DARN fuses the feature maps and adds a tree-structured layer on the top for category and attribute prediction layer. FashionNet~\cite{Deepfashion} detects garment landmarks from various scenarios where results are used to subsample the feature maps of the last convolutional layer, considerably suppressing background noise and disturbance. Nevertheless, the above-mentioned methods directly concatenate outputs from different layers to form a query vector, unavoidably resulting in high computational cost during query procedure. In contrast, our method follows deep hashing fashion which learns discrete embeddings for samples and accelerates querying up to $306\times$ without compromising the accuracy.

\subsection{Attribute Learning}
Researches on attributed-based visual representation have received wide attention by the computer vision community in the past few years, especially in person re-identification~\cite{person1, person2}, captioning~\cite{caption1, caption2, caption3}, and retrieval~\cite{ret1, ret2, recipe}. Attributes are usually referred as semantic properties of the objects or the scenes that are shared across categories so that attributes could serve as a latent and interpretable connection between image content and abstract labels. Previous work~\cite{DARN,Deepfashion} purely views attribute learning as a multi-label classification problem with a global representation, yet its performance limits especially when confronted with excessive number of fine-grained attributes. \cite{concept} learns the model for automatically grouping garment attributes into an upper-level concept list (\textit{e.g.,} the neckline concept might consist of attributes like v-neck, round-neck) while \cite{DARN} constructs a tree structure for attribute hierarchy. However, a fixed tree-structure design is not flexible enough for learning emerging new attributes. From the linguistic point of view, people describe a noun with a set of adjectives conforming to a certain order. Exploiting a sequence of attributes rather than isolated attributes could preserve a stronger sequential connection between attributes, like ``floral'' is usually after ``bohemian''. Therefore, in order to further explore the word order, we derive the decoder part in the proposed model for generating attribute descriptions in the light of fast development of Long Short Term Memory networks~(LSTM)~\cite{LSTM}.

\subsection{Multi-task Learning}
Multi-task learning~(MTL) aims to improve generalization performance of multiple prediction tasks by appropriately sharing relevant information across them. In the context of deep neural networks, this idea is often realized by hand-designed network architectures with layers that are shared across tasks and branches that encode task-specific features~\cite{DARN,3dpoint}. \cite{CP-mtML} integrates multiple face matching criteria to transform images into task-specific subspace which assists to learn a common projection for metric-balanced face retrieval. \cite{DPH,DARN,Deepfashion} jointly preserve the category
and attribute similarities for image retrieval by applying cross-entropy objective function for attribute prediction and softmax loss for classification. The experiments we conducted also show that multi-task learning greatly leverages on both related tasks by sharing task-specific knowledge.

\section{Our Approach}
In this paper, we propose a joint framework to perform the tasks (1) sequential attribute learning; and (2) the cross-domain garment image embedding simultaneously. After presenting the problem statement, we give detailed descriptions of the generic neural encoder-decoder framework for sequential attribute learning, followed by the explanation of the proposed attention-based cross-domain embedding.

\subsection{Problem Statement}

We denote the training image from the user domain and its corresponding shop image (\textit{i.e.,} positive sample) and irrelevant shop image (\textit{i.e.,} negative sample) from database images as $I$, $I_p$ and $I_q$ respectively. One objective of the proposed model is to generate a T-gram description of the image $I$, $y = \{ y_1, y_2, \cdots, y_{T}\}$. In the meanwhile, the model jointly 
encodes images $I$, $I_p$, and $I_q$ as a set of $C$-length binary codes $e$, $e_{I_p}$, and $e_{I_q}$ $\in$ $\{-1, +1\}^{1\times C}$ respectively, by which the embeddings of positive user-shop image pairs are likely to be similar or close in the projected Hamming space.

\subsection{Encoder-Decoder for Sequential Attribute Learning}
Given an image and its corresponding attribute descriptions, the encoder-decoder model directly maximizes the following objective:

\begin{equation}
\theta^* = \max_{\theta}\sum_{I,y}\log p(y|I;\theta),
\end{equation}
where $\theta$ are the parameters of the model, $I$ denotes the image and $y = \{y_1, y_2 \dots, y_{T}\}$ is the corresponding attribute description, consisting of $T$ words. Using the chain rule, the \emph{log} likelihood of the joint probability distribution can be decomposed into the ordered conditionals:
\begin{equation}
\log p(y) = \sum_{t=1}^T \log p(y_t|y_1, y_2, \dots, y_{t-1}, I),
\end{equation}
where we drop the dependency on model parameters for convenience. It is natural to model $p(y_t|y_1, y_2, \cdots, y_{t-1}, I)$ with a recurrent neural network~(RNN), where the variable number of words we condition upon to  time $t-1$ is expressed by a fixed length hidden state or memory $h_t$. Hence as \eqref{eq:lstm} shows, we adopt a nonlinear function to model the transform from $h_{t-1}$ to $h_t$. To make the RNN valid and effective, it is crucial to decide: what is the exact form of $f$ and how are the images and words fed as inputs $x_t$ at time t. 
\begin{equation}
h_{t} = f(h_{t-1}, c_{t-1}) = LSTM(h_{t-1}, x_{t-1}, m_{t-1}).
\label{eq:lstm}
\end{equation}
For $f$ we use Long-Short Term Memory~(LSTM) net, which is shown powerful performance on sequence tasks. Here $c_{t-1}$ represents for context vector, $x_{t-1}$ is the input vector and $m_{t-1}$ the memory cell vector at time $t-1$ respectively. In conventional encoder-decoder settings, the context vector $c_{t}$ only depends on the output of the encoder, a Convolutional Neural Network~(CNN). The output features extracted from the last fully connected layer give the global visual information of the input image. During the decoder stage, $c_t$ usually keeps constant.

Things are different in the attention-based framework. $c_t$ will be updated during the whole training procedure because the focused part of the image is changing as the predicted words shift. To compute the context vector $c_t$, inspired by \cite{adaptive}, we form a spatial attention model, which is defined as \eqref{eq:ct}.
\begin{equation}
c_t = g(V, h_t),
\label{eq:ct}
\end{equation}
where $g(\cdot)$ is an attention function, and $V = [v_1, \dots, v_K], v_i \in \mathbb{R}^{D\times 1}$ is a $D$ dimensional spatial feature vector corresponding to the $i$-th region. 

Given the spatial image feature $V \in \mathbb{R}^{D\times K}$ and the hidden state $h_t \in\mathbb{R}^{D\times 1}$ of the LSTM, we feed them through a single layer neural network followed by a softmax function to generate the attention distribution over the $K$ regions of the images.

\begin{equation}
\begin{split}
z_t &= w_h^T\tanh(W_vV + (W_gh_t)\mathbb{1}^T), \\
\alpha_t &= softmax(z_t),
\end{split}
\label{eq:zt}
\end{equation}
where $\mathbb{1}\in \mathbb{R}^K$ is a vector with all elements of 1. $W_v, W_g \in \mathbb{R}^{K\times D}$ and $w_h\in \mathbb{R}^{K}$ are parameters to be learned. $\alpha_t\in \mathbb{R}^K$ is the attention weight over features in $V$. Therefore, the context vector $c_t$ can be obtained by a weighted sum:  
\begin{equation}
c_t = \sum_{i=1}^K\alpha_{ti}v_{i}.
\end{equation}
So far, we have obtained a location-aware visual feature that allows the decoder to "speak" from different views. Yet the spatial attention based decoders still cannot determine when to rely on visual signal and when to rely on the language model. We adopt the visual sentinel, which is a gate that allows decoder to choose whether to focus on linguistic rules of attribute description or the image visual content. 
\begin{equation}
\begin{split}
g_t &= \sigma(W_xx_t + W_hh_{t-1}),\\
s_t &= g_t\odot\tanh(m_t),
\end{split}
\end{equation}
where $W_x$ and $W_h$ are weight parameters to be learned, and $g_t$ is the gate applied to the memory cell $m_t$. $\odot$ represents the element-wise product and $\sigma$ is the logistic sigmoid activation. 

Therefore, the new context vector, defined as $\hat{c_t}$ could be calculated as follows,
\begin{equation}
\hat{c_t} = \beta_ts_t + (1-\beta_t)c_t,
\end{equation}
where $\beta_t$ is the new sentinel gate at time $t$ and it produces a scalar in the range [0,1]. The scalar helps to balance the importance of the visual sentinel information and the image spatial attention. To compute $\beta_t$, we modify the equation \eqref{eq:zt},
\begin{equation}
\begin{split}
\hat{\alpha} &= softmax([z_t; w_h^T\tanh(W_ss_t + W_gh_t)]),\\
\beta_t &= \hat{\alpha_t}[K+1].
\end{split}
\end{equation}
$\hat{\alpha}\in\mathbb{R}^{K+1}$ is the attention distribution over both spatial features and the visual sentinel vector.

The probability over a set of possible attributes at time $t$ can be calculated as
\begin{equation}
p_t(y_t) = softmax(W_p(\hat{c_t} + h_t)),
\end{equation}
where $W_p$ is the weight parameter to be learned. Hence the loss function of attribute learning part is
\begin{equation}
L_{tag}(I,y) = -\sum_{t=1}^T\log p_t(y_t)
\label{eq:loss1}
\end{equation}

\begin{figure}[!htb]
	\centering
	\includegraphics[width=1\linewidth]{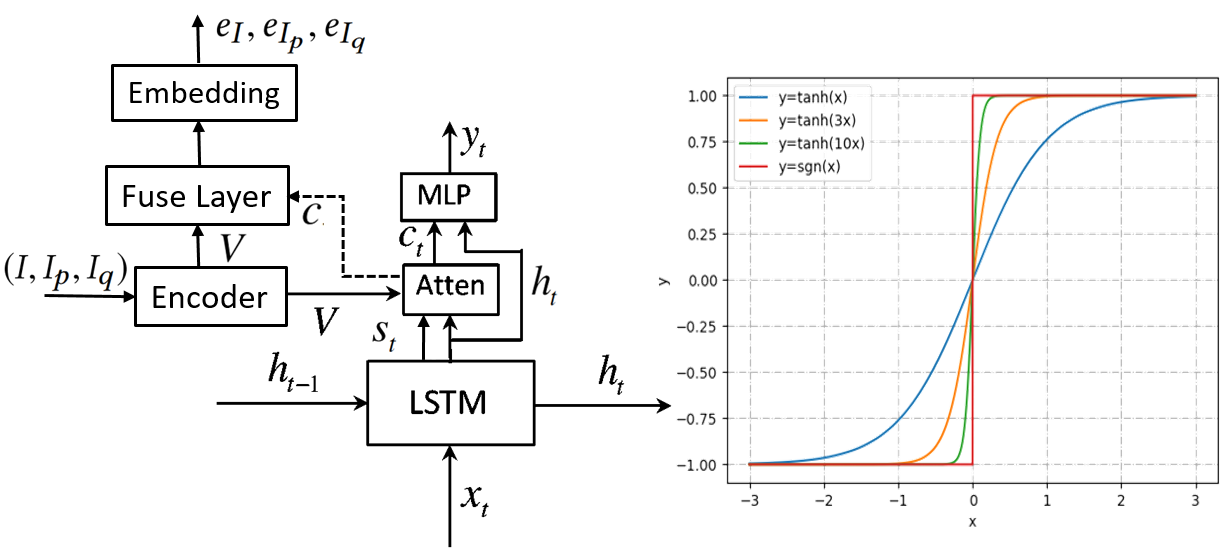}\\
	\caption{(Left) An illustration of our proposed encoder-decoder structure. The dotted lines denotes sum of output of attention modules. (Right) An illustration of the process through which $tanh(\cdot)$ approximates $sgn(\cdot)$ gradually.}
	\label{fig:LSTM}
\end{figure}

\subsection{Attention-aware Embedding for Cross-domain Retrieval}
In this part, we will elaborate how the sequential attribute learning would help image to embed. Previously, the binary hash code is learned directly by converting $\tau$-length feature maps $z$ from the last fully-connected layer, which is continuous in nature, to a binary code $b$ taking values of either $+1$ or $-1$. This binarization process can only be realized by taking the signum function $b = sgn(z)$ as the activation function on the top of the embedding layer. 
\begin{equation}
b=sgn(z) = \left\{
\begin{aligned}
+1, ~&\text{if}~ z\geq 0 \\
-1,  ~&\text{otherwise}
\end{aligned}
\right.
\end{equation}
Unfortunately, as the signum function is non-smooth and non-convex, its gradient is zero for all nonzero inputs, and is ill-defined at zeros, which makes the standard back-propagation infeasible for training deep networks. This is known as the vanishing gradient problem, which has been a key difficulty in training deep neural networks via back-propagation\cite{deepresidual,bn,dropout}. Approximation solutions that relax the binary constraints are not good alternativse as they lead to a large quantization error and therefore to a suboptimal performance. 

In order to alleviate the optimizing problem of non-smooth signum activation, we draw inspiration from recent studies on continuation methods~\cite{hashnet, jingkuan}. These studies propose a strategy by gradually reducing the amount of smoothing during the training, which results in a sequence of optimization problems eventually converging to the original optimization problem. Following this strategy, if we find an approximate smooth function of $sgn(\cdot)$, and then progressively make the smoothed objective function non-smooth as the training proceeds, the final solution should converge to the desired optimization target.

Motivated by the continuation methods, the $tanh(\cdot)$ function is applied to approximate $sgn(\cdot)$. It is also noticed that there exists a critical relationship between the $sgn(\cdot)$ and the $tanh(\cdot)$. As illustrated in Figure \ref{fig:LSTM}, increasing scaling parameter $\phi$, the scaled tanh function will become more non-smooth and more satuated, so that the deep networks using $\tanh(\phi z)$ as the activation function will be more difficult to optimize. As $\phi\rightarrow\infty$, the optimization problem will converge to the original deep hashing problem with $sgn(z)$ activation function:

\begin{equation}
sgn(z) = \lim_{\phi\rightarrow\infty}\tanh(\phi z).
\end{equation}

In order to highlight local details of garment images, we fuse the global image feature $V$ with weighted local features and use $\tanh(\cdot)$ to approximate discretization,
\begin{equation}
\begin{split}
z &= W_e([V;\lambda\frac{1}{T} \sum_{t=1}^Tc_t]),\\
e_I &= \tanh(\phi z), s.t. \phi\rightarrow\infty
\end{split}
\end{equation}
where $W_e$ is the parameter to be learned, $\phi$ is a scaling vector being gradually enlarged after each time the network converges, and $[\cdot;\cdot]$ denotes concatenation. $\lambda$ is the weight to balance the global features and the attended features. As discussed in Section 1, the cross-domain retrieval involves thousands fine-grained classes, which is not suitable to be applied a simple cross-entropy loss. In this case, we adopt the triplet loss function as the training objectives for embedding, which is shown as below.
\begin{equation}
L_{emb}(I, I_p, I_q) = \max(0, d(e_I, e_{I_p}) - d(e_I, e_{I_q}) + \gamma),
\label{eq:loss2}
\end{equation}
where $I$ is the anchor image from the user domain, $I_p$/ $I_q$ is the positive/negative image from the shop domain. $\gamma$ is a constant for setting the margin and $d(\cdot)$ measures the distance of two image embeddings. The loss function penalizes the triplets if $d(e_I, e_{I_p}) + \gamma > d(e_I, e_{I_q})$ to make matched image embeddings close and unmatched image embeddings distant in the common space. Thus, the multi-task loss is combined with \eqref{eq:loss1} and \eqref{eq:loss2},
\begin{equation}
L = L_{tag}(I) + L_{tag}(I_p) + L_{tag}(I_q) + \eta L_{emb}(I, I_p, I_q),
\end{equation}
where $\eta$ is a weight, which is empirically set to $3$, in case that $L_{tag}$ dominates the training.

\section{Experiments}
In this section, we evaluate our DMCH on both tasks of cross-domain retrieval and sequential attribute learning on two large-scale fashion datasets. 
\subsection{Datasets}

\subsubsection{\textbf{DARN}\cite{DARN}} This dataset is created for street-to-shop retrieval, \textit{i.e.,} matching street images taken by users with professional shop photos. After removing corrupted images, a subset of 62,812 street images and 39,756 shop images over 20 categories are generated. Each street image has a matched shop image. The dataset also provides a sequence of attributes for each clothing item. We follow the settings in \cite{Attention} and generate 2,076,440 training triplets in the form of  $($\emph{user image}, \emph{positive sample}, \emph{negative sample}$)$ with the positive-to-negative ratio of 1-to-10. A number of 2,000 distinct user images are randomly selected for testing, each of which corresponds to a unique clothing item. A total of 102 attributes are involved under this setting.

\subsubsection{\textbf{DeepFashion}\cite{Deepfashion}} This dataset includes 105,562 user images and 28,512 shop images of 19,135 unique clothing items. Additionally, the auxiliary data including category labels, clothing attributes, clothing landmarks, and street-shop image pairs are also provided. Each clothing item may have a set of street images but only a couple of shop images. 1,911,570 training triplets with the positive-to-negative ratio of 1-to-10 are generated without overlap of test dataset. 237 relevant attributes are chosen for taxonomy of attribute learning. 4,582 street images are randomly selected as testing queries to search against the entire 28,512 shop images for evaluation of cross-domain task. Different from [21], landmark data is not used in our experiments as it is not required by the proposed model.

\subsection{Experimental Setting}
All the experiments are conducted on a server with Intel Xeon(R) CPU E5-2660 and two Telsa K40c GPU cards. The basic model applied for encoder is ResNet-152~\cite{deepresidual}, on which the last two layers are removed for fine-tuning. Mini-batch Stochastic Gradient is used for parameter updating. The hidden size is fixed at 256 and the dimensionality of word embedding vectors is 256. The batch size is fixed at 32 and the momentum is 0.9. The learning rate is set to 0.001 and decays by 0.1 for every 50 epochs. Margin in the triplet loss function is set to $\frac{1}{16}$ of the hash code length (\textit{e.g.,} 2 for 32-bit hash codes) for the experiments. The scaling parameters $W_e$ is enlarged by 10 times after each time the network converges. 
\subsubsection{Evaluation Metric}
Following \cite{Attention}, we evaluate retrieval performance by the top-$K$ precision, which is defined as follows:
\begin{equation}
P@K = \frac{\sum_{q\in Q}hit(q,K)}{|Q|},
\end{equation}
where Q is the total number of queries; $hit(q,K)=1$ if at least one image of the same product as the query image $q$ appears in the returned top-K ranking list; otherwise $hit(q,k)=0$. For most queries, there is only one matched shop image in both the DARN and DeepFashion datasets.

For evaluation of sequential attribute learning, we employ BLEU, ROUGE-L, and CIDEr metrics. BLEU computes the geometric mean of the n-gram precisions,
\begin{equation}
BLEU_n = \frac{\sum_{gram_n\in pred}hit(gram_n, gt)}{\sum_{gram_n\in pred}|ngram|},
\end{equation}
where $gt$ and $pred$ denote the ground-truth and the prediction of $n$-length attribute sequences, respectively. Similarly, ROUGE-L~\cite{rouge} is a recall-oriented measure that evaluates the quality of the longest common subsequence and CIDEr~\cite{cider} measures the overall quality of the generated attribute sequence against the ground truth provided by humans.

\subsubsection{Compared Methods}
To achieve a fair comparison on accuracy and efficiency of cross-domain garment image retrieval, we compare the proposed model with the state-of-the-art methods in the literature such as WTBI~\cite{Wtbi}, DARN~\cite{DARN}, FashionNet~\cite{Deepfashion}, TagYNIN and CtxNIN~\cite{Attention}. Note that the length of feature vector varies from 17,920-D (4,096-D after PCA)~\cite{Deepfashion,DARN,Wtbi} to 1024-D~\cite{Attention}. Additionally, we juxtapose deep hashing with DMCH with a fixed embedding length (\textit{i.e.,} 128-bit) to observe whether the defined multi-task objectives contribute positively.

\subsection{Evaluation on Cross-domain Retrieval}
\subsubsection{Comparisons with the State-of-the-art Methods}
\begin{figure}[htb!]
	\centering
	\includegraphics[width=0.45\textwidth]{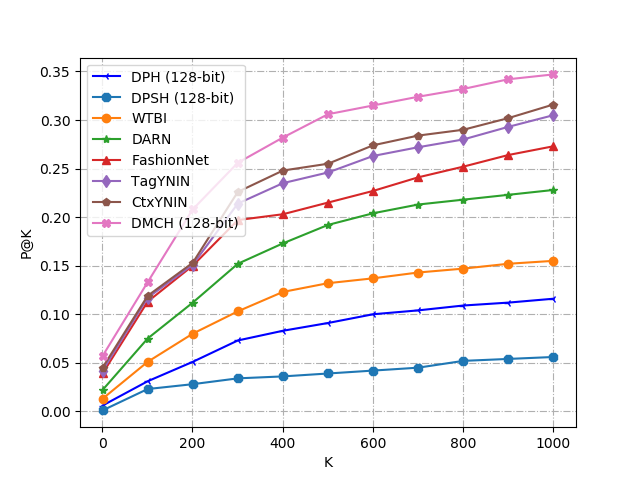}
	\caption{Comparison of P@K on the DeepFashion Dataset. }
	\label{fig:Deepfashion}
\end{figure}
Figure \ref{fig:Deepfashion} shows the top-K precision of the baseline methods and our mechanism on the DeepFashion Dataset. We can clearly see that the WTBI is inferior to other cross-domain retrieval methods since they only apply attributes to expand the categories rather than exploiting their correlations. DARN performs better than WTBI as it learns different branches of NIN~(Network in Network~\cite{nin}) for user domain and shop domain and employs a tree-structured layer for attribute prediction, which potentially guides the model to reach strong discrimination. However, it is not competitive to FashionNet, which further applies clothes detection as an initialization step to suppress the noise from image background. FashionNet shares the convolutional layers for both domains while uses different top branches for two tasks (\textit{i.e.,} including attribute prediction and landmark prediction). Though YNet similarly locates the major attention of image beforehand, TNet chooes to utilize attribute information as input and constructs two subnetworks for separate domains. The representation of user images in YNet is endowed strong connection with both positive and negative shop image representations. In this case, YNet shows superior performance among the aforementioned methods, however, it uses attributes as inputs to explore garment attention, which makes it uncommercial for practical application. As for deep hashing based methodologies, We fix the code length to 128-bit for fair comparisons. As we can see, DPH performs better than DPSH since DPH adopts the attribute-preserving strategy, which potentially assists to adjust embeddings for fine-grained categories. However, as discussed in Section 2.1, they are not specifically designed for cross-domain task, where disturbances and noises from the image background exist. Our DMCH modal consistently achieves the best performance on the cross-domain garment image retrieval task, mainly owing to the attention mechanism for local details recognition, sequential modeling for attributes and iterative optimization for embedding. 

\begin{figure}[htb!]
	\centering
	\includegraphics[width=0.45\textwidth]{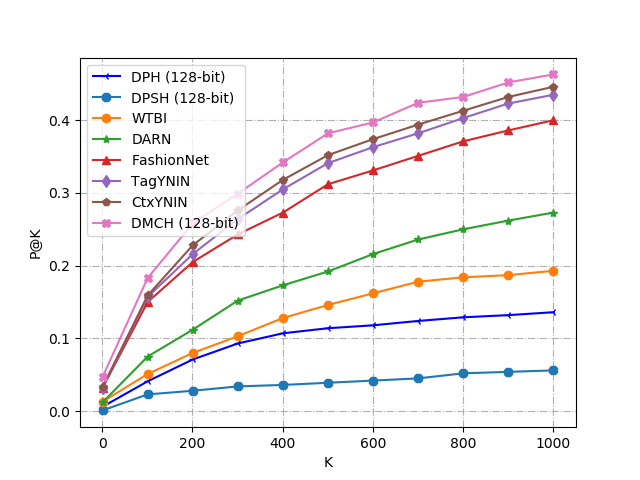}
	\caption{Comparison of P@K on the DARN dataset. }
	\label{fig:DARN}
\end{figure}
Figure \ref{fig:DARN} shows the top-K precision of the baseline methods and our mechanism on the DARN dataset. The proposed method still remains relatively superior, while the gap between other attribute based methods is closer than that from the DeepFashion dataset. The reason of the phenomenon could be that, in the DARN dataset, each item is associated with attributes of shorter length, makes sequential attribute learning slightly degrade to conventional attribute classification. Nevertheless, the ground-truth of attributes in both DeepFashion and DARN datasets are still incomplete and vague (\textit{e.g.,} ``thick'' vs ``regular thickness''). Even human annotators are easy to misjudge these words. A better performance is expected by refining the attributes before learning.  
\begin{table}[!htb]
	\centering
	\caption{The Precision@200~(\%) of DMCH on DeepFashion and DARN using different combinations of loss with 128-bit code.}
	\label{tab:loss}
	\resizebox{0.5\textwidth}{!}{%
	\begin{tabular}{l|l|ccc}
		\cline{1-3}
		Components  & DeepFashion & DARN &  &  \\ \cline{1-3}
		$L_{embed}$ &    2.1         & 2.8     &  &  \\ \cline{1-3}
		$L_{embed}$ + $L_{attr}$          &  13.7           &19.6    &  &  \\ \cline{1-3}
		$L_{embed}$ + $L_{tag}$            & \textbf{20.8}            & \textbf{25.8}    &  &  \\ \cline{1-3}
	\end{tabular}
}
\end{table}

\subsubsection{Analysis on Multi-task Loss}
In this subsection, we study the effect of loss componenst for cross-domain retrieval task on the DeepFashion dataset with a fixed code length (128-bit). The loss function designed for DMCH mainly consists of the triplet embedding loss ($L_{embed}$) and the cross-entropy loss ($L_{tag}$). The conventional attribute learning is positioned as a multi-label classification problem, which could be learned with a fully connected layer supervised by the cross-entropy loss ($L_{attr}$). We report the retrieval result with respect to different sorts of loss functions in Table \ref{tab:loss}. The observation is two-fold, 1) attribute supervision contributes positively to bridging fine-grained classes and endowering embeddings with rich inpretable semantics; 2) taking attribute order into consideration, the proposed sequential attribute learning improves the top-200 precision by up to 51.8\% relatively, compared with conventional attribute learning. Usually when classes number goes up, the ``long-tail'' distribution problem on classification emerges, that is, rare attributes from the tail of distribution becomes much harder to predict. However, our sequential attribute learning exploites conditioned probability to adjust the attribute distribution at every time $t$, which potentially alleviates the ``long-tail'' problem.
\begin{table}
	\caption{The Precision@200~(\%) of DMCH on the DeepFashion and DARN dataset. We compare the model with various numbers of bits and with different discretization strategies.}
	\resizebox{0.5\textwidth}{!}{%
	\begin{tabular}{l|ccc|ccc}
		\toprule
		\multirow{2}{*}{Methods} &
		\multicolumn{3}{c}{DeepFashion} &
		\multicolumn{3}{c}{DARN} \\
		& {32-bit} & {64-bit} & {128-bit}  & {32-bit} & {64-bit} & {128-bit} \\
		\midrule
		two-stage solution &6.8 &11.5 &17.1 &5.6 &12.6 &20.3 \\
		$\tanh(z)$ &8.6 & 13.2 &19.1 &7.6 &14.4 &22.2 \\
		$\lim_{\phi\rightarrow\infty}\tanh(\phi z)$ &10.1 &14.6 &\textbf{20.8} &9.8 &16.7 &\textbf{25.8} \\
		\bottomrule
	\end{tabular}
}
	\label{hashing}
\end{table}

\subsubsection{Study on Binary Optimization and Code Length}
The proposed DMCH is capable of generating hash codes directly while most of previous hashing methods follow the two-stage strategy, \textit{i.e.,} first to learn continuous representations and then do discretization with a non-smooth signum function. In this subsection, we primarily investigate the effect of direct binary codes optimization and code length of embedding on retrieval performance, which is shown in Table \ref{hashing}. It is observed that the post-processing signum function and the vanilla tanh function are both sub-optimal as they suffer the quantization error with different levels. The proposed iterative optimization alleviates the quantization error of hashing, and shows superior performance among other discretization strategies. As for the code length, we could observe that longer embedding enables to preserve more visual and semantic information thus leading to a better retrieval performance.

\subsubsection{Study on Querying Efficiency of DMCH}
\begin{figure}[htb!]
	\centering
	\includegraphics[width=0.45\textwidth]{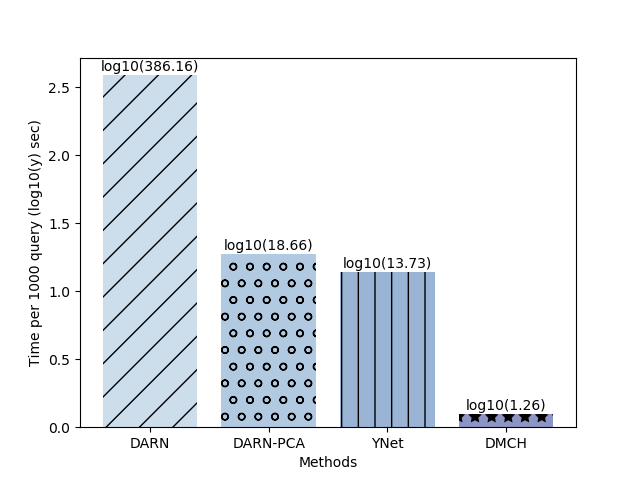}
	\caption{Running time for 1,000 query samples with different cross-domain retrieval methods on the DARN dataset. Note feature vectors or embeddings are of various length.}
	\label{fig:efficiency}
\end{figure}

To study the efficiency of the proposed model, we measure the running time of 1,000 query samples on the DARN dataset, which is shown in Figure \ref{fig:efficiency}. DARN and FashionNet concatenate the local features from the convolutional layers and the global features from the fully connected layers (\textit{i.e.,} 17,920-D). Consequently, their feature dimensionality is much larger than that of YNet~\cite{Attention} (\textit{i.e.,} 1000-d for NIN structure). Even after dimensionality reduction by PCA, the feature dimensionality of DARN is still up to 4096-D. In contrast, the proposed method jointly embeds images into binary codes with much shorter length (\textit{e.g.,} 128-bit), as well query processing is significantly accelerated as the Hamming distances are calculated instead of expensive Euclidean distance calculation. It is observed that our approach speeds up $10.9\times$, $14.8\times$, $306.5\times$ in comparison with Y-Net, DARN-PCA and DARN respectively.

\begin{figure}[htb!]
	\centering
	\includegraphics[width=0.45\textwidth]{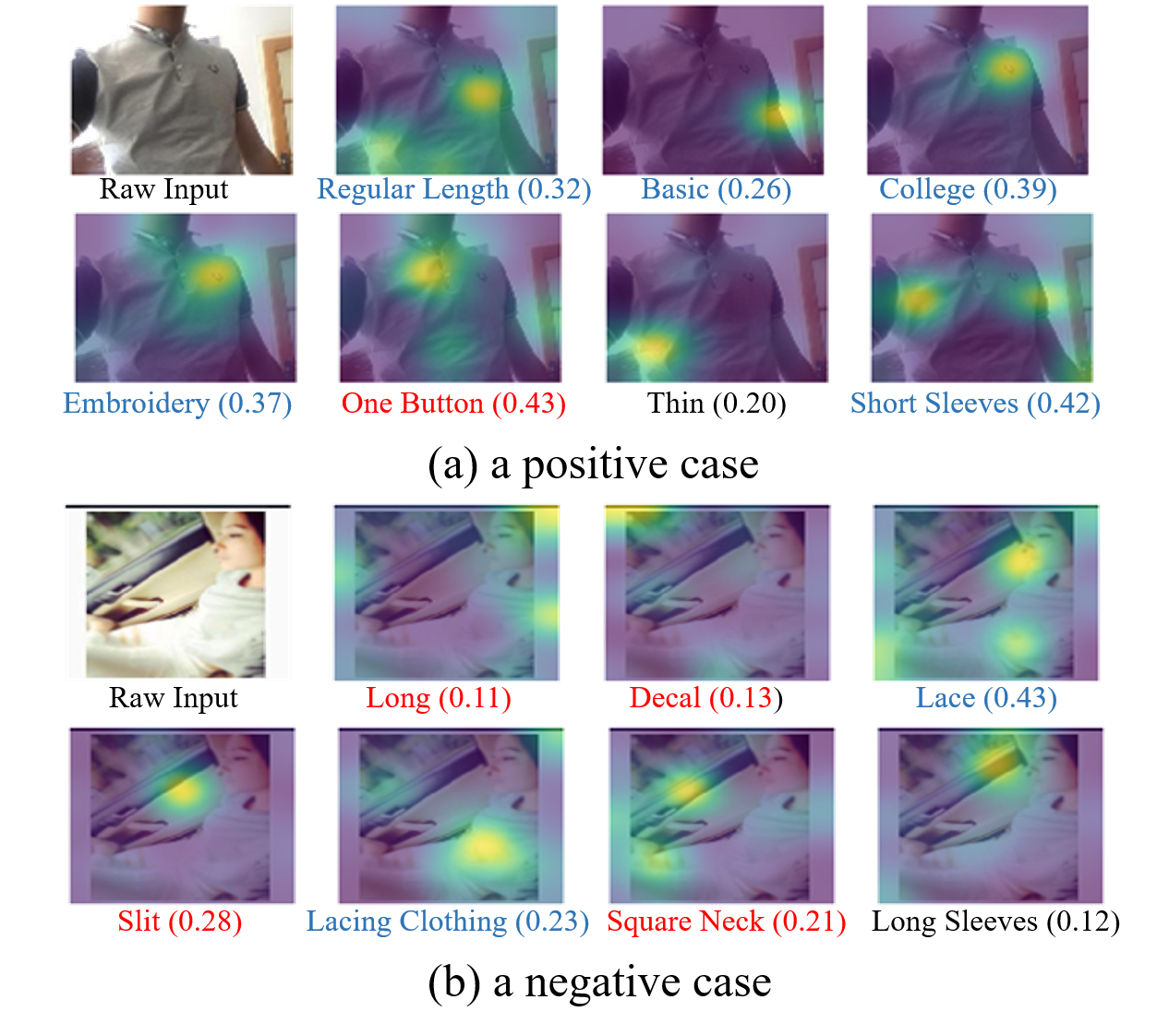}
	\caption{Visualization of the spatial attention maps. Correct attribute prediction is shown in blue captions, wrong ones in red captions, and unknown ones in black captions.}
	\label{fig:attention}
\end{figure}

\begin{table*}
	\caption{The performance of sequential attribute learning on the DeepFashion and DARN dataset. The code length of DMCH is fixed at 128-bit.}
	\resizebox{1\textwidth}{!}{%
	\begin{tabular}{lcccccccccccc}
		\toprule
		\multirow{2}{*}{Methods} &
		\multicolumn{6}{c}{DeepFashion} &
		\multicolumn{6}{c}{DARN} \\
		& {B-1} & {B-2} & {B-3} & {B-4} & {ROUGE-L} & {CIDEr} & {B-1} & {B-2} & {B-3} & {B-4} & {ROUGE-L} & {CIDEr}\\
		\midrule
		WTBI &14.1 &11.4 &7.1 &5.5 &16.7 &13.6 &17.3 &14.2 &10.1 &7.6 &19.6 &24.6\\
		DARN &28.5 & 19.6 &11.6 &7.6 &33.4 &24.2 &34.6 &24.7 &15.8 &9.3 &32.5 &53.2\\
		FashionNet &\textbf{36.8} &21.3 &13.7 &9.8 &35.8 &26.4 &42.3 &26.6 &18.7 &11.5 &42.0 &76.5\\
		DMCH &35.7 &\textbf{25.9} &\textbf{19.1} &\textbf{14.2}  &\textbf{44.1} &\textbf{31.5} &\textbf{47.7} &\textbf{34.2} &\textbf{25.7} &\textbf{18.5} &\textbf{51.0} &\textbf{98.6}\\
		\bottomrule
	\end{tabular}
	}
	\label{tab:tag}
\end{table*}

\begin{figure*}[htb!]
	\centering        \includegraphics[width=0.85\textwidth]{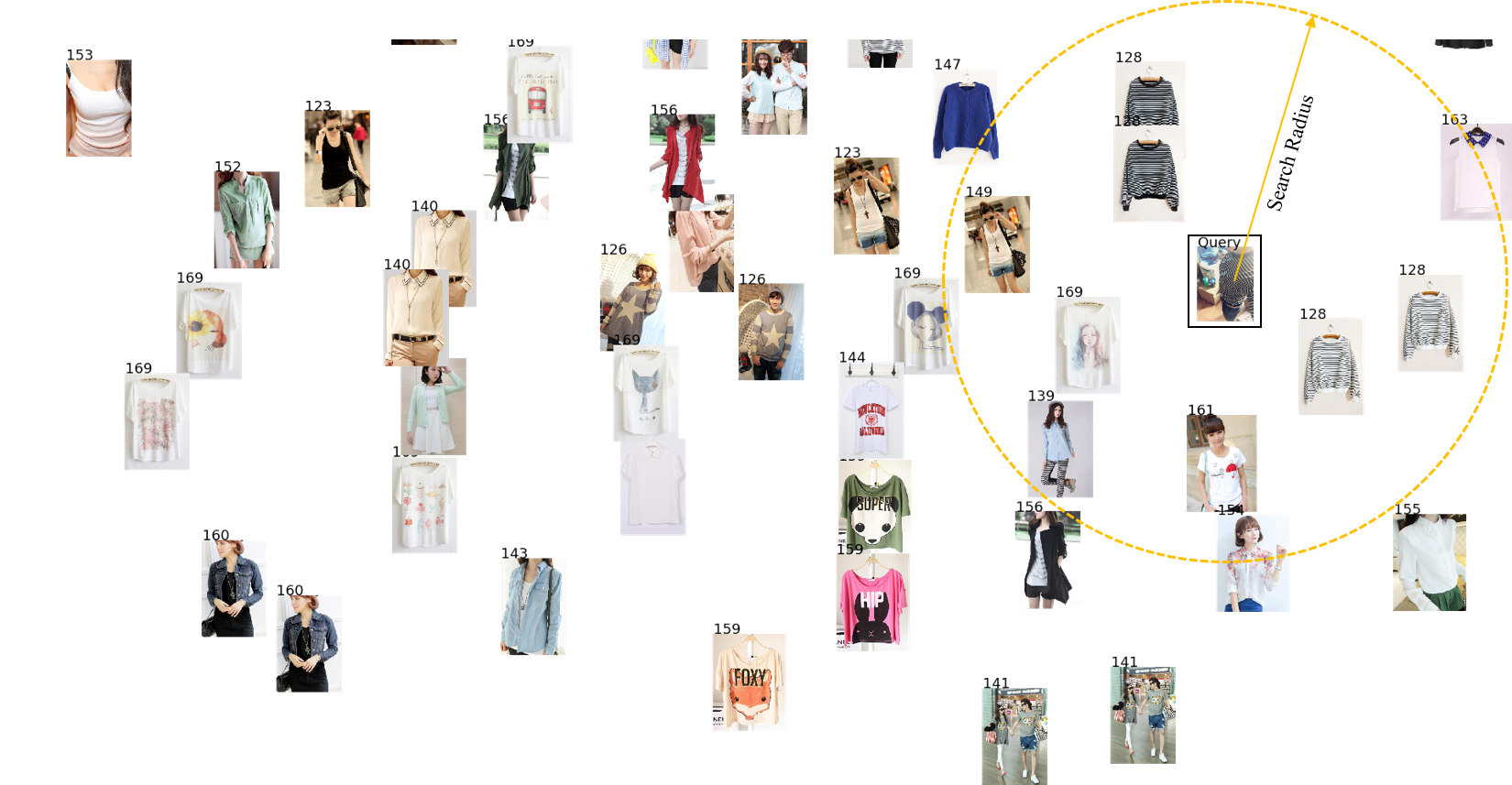}
	\caption{A 2D visualization of the embedding Hamming space of cross-domain images. A query user image is bounded by a black box, and the yellow arrow represents the pre-defined search radius. The shop images falling in the yellow circle will be returned as the matched clothing items of the query image. Each shop image is associated with a product ID. A clothing item may have more than one shop images.}
	\label{fig:ret}
\end{figure*}

\subsection{Evaluation on Sequential Attribute Learning}
\subsubsection{Comparison with State-of-the-art Methods}
In this part, we mainly target at verifying the accuracy and comprehensiveness of our sequential attribute learning task. A detailed comparison among different methods is illustrated in Table \ref{tab:tag}. Attributes defined in both datasets follow a certain order similar to human being linguistic convention. The WTBI achieves the relatively lowest scores across all metrics, while DARN performs averagely better. Besides, FashionNet is quite competitive to the proposed method. 
In terms of the unary metric (\textit{i.e.,} BLEU-1), FashionNet achieves a slight gain over DMCH on the DeepFashion Dataset. It is natural to see the multi-label classification model hits accurately on single words as it is not required to consider the context and the order. Thus thinking through the context metric (\textit{e.g.,} BLEU-2~4, ROUGE-L, CIDEr), our model shows much better performance since the algorithm chooses to memorize the correlation of sequential data besides isolated attributes. It is also noticed that the performance on the DARN dataset is much better than the result on the DeepFashion. That is potentially because the DARN dataset contains much shorter attributes among a small set of taxonomy compared with the DeepFashion Dataset.

\subsubsection{Attention Visualization}
For better understanding of the validity of our attention modules, we visualize the attention maps of two sample images from the DeepFashion dataset in Figure \ref{fig:attention}. The capability of the proposed model on attribute learning is clearly demonstrated, including discovering the spatial correlation (\textit{e.g.,} two attentive parts for ``short sleeves'') and preserving the semantic correlation (\textit{e.g.,} ``lace'' and ``lacing clothing''). Take our sequential attribute learning result in Figure \ref{fig:attention} as an example. The first clothing item is described in the order of ``length'', ``details and style of clothes'', ``thickness of clothes'', and ``collar and sleeve type" where the linguistic order of descriptors is well reserved. A failure case shown in the second example is caused by an incomplete view of the clothing item and its confusing color to the background.

\subsection{Exemplars of Cross-domain Search}
In this subsection, we use real garment images from the DeepFashion dataset to visualize the leant embedding space with t-SNE~\cite{t-sne} in Figure \ref{fig:ret}, which gives an intuitive understanding on the cross-domain task and the performance. By projecting 128-bit binary codes to a 2D plane, three potential issues with the training dataset can be clearly observed: 1) highly similar shop images are used by different online merchants. For instance, both items ``123'' and ``149'' use a same shop image; 2) the shop images labeled with a same product ID are of large variance. Take items ``169'' and ``159'' as an example, the major patterns, the color, and the text on the clothing items are quite different; 3) the categories of products are numerous. It is nearly impossible to get all mutual relationships involved in our training triplets. Better performance of cross-domain garment image retrieval is expected by addressing the above data quality issues in future.

\section{Conclusion and Future Work}
To deal with the problem of cross-domain garment image search, we have proposed a novel joint learning framework which shares visual and verbal knowledge to exploit relationship among fine-grained classes and attributes. Different from the state-of-the-art models in the literature, we treat attribute descriptions as rich context for activating spatial attention that enhances recognition of garment details. Meanwhile we embed images as binary codes with short length to significantly improves the query efficiency. Since our model is of great potential, we will further explore multi-modal retrieval in the near future.
\newpage


%





\ifCLASSOPTIONcaptionsoff
  \newpage
\fi





\bibliographystyle{IEEEtran}
\bibliography{MTT_reveyrand}
%

\begin{IEEEbiography}[{\includegraphics[width=1in,height=1.25in,clip,keepaspectratio]{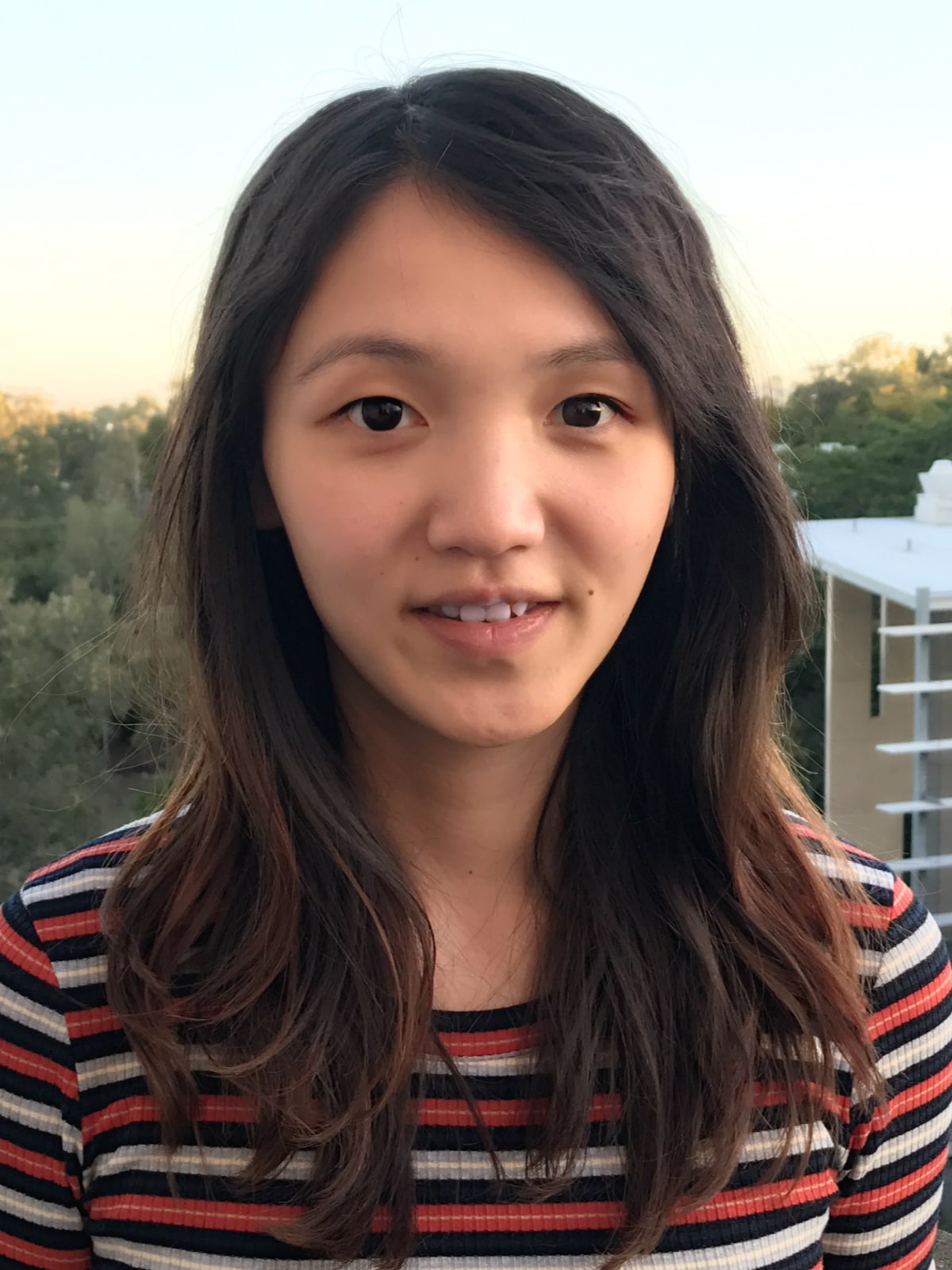}}]{Yadan Luo}
received the B.S. degree in computer science from the University of Electronic Engineering and Technology of China in 2017, and is currently working toward the Ph.D. degree at the University of Queensland. Her research interests include multimedia retrieval, machine learning and computer vision.
\end{IEEEbiography}

\begin{IEEEbiography}[{\includegraphics[width=1in,height=1.25in,clip,keepaspectratio]{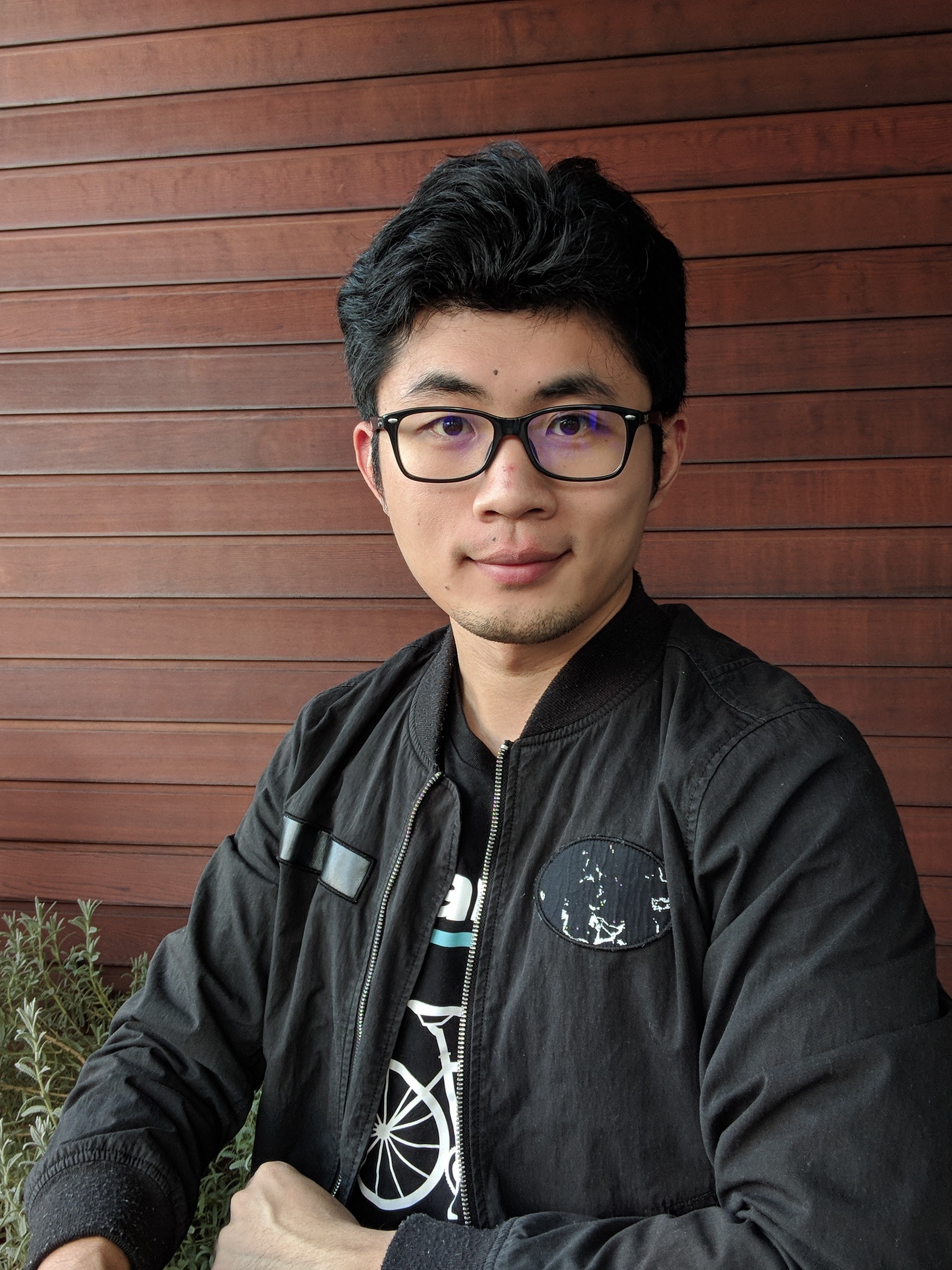}}]{Ziwei Wang} received his BSc degree from Beijing University of Civil Engineering and Architecture in 2014 and his Master degree of Computer Science from The University of Queensland, Australia in 2016. He is currently a PhD candidate at The University of Queensland. His research interests include image captioning and machine learning. 
\end{IEEEbiography}

\begin{IEEEbiography}[{\includegraphics[width=1in,height=1.25in,clip,keepaspectratio]{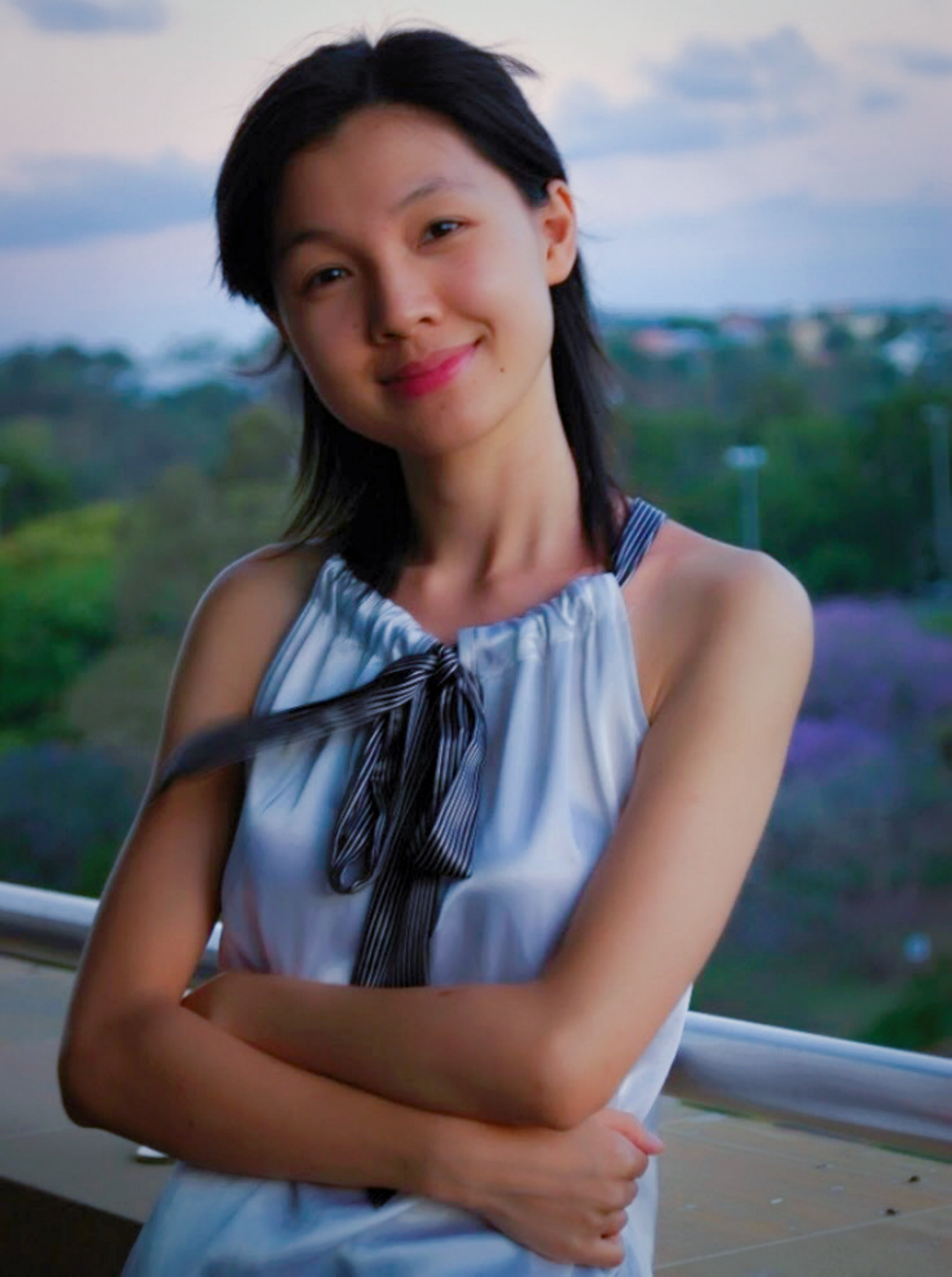}}]{Zi Huang}
is an ARC Future Fellow in School of ITEE, The University
of Queensland. She received her BSc degree from Department of Computer Science, Tsinghua University, China, and her PhD in Computer Science
from School of ITEE, The University of Queensland. Dr. Huang's research
interests mainly include multimedia indexing and search, social data analysis and knowledge discovery.
\end{IEEEbiography}
 
\begin{IEEEbiography}[{\includegraphics[width=1in,height=1.25in,clip,keepaspectratio]{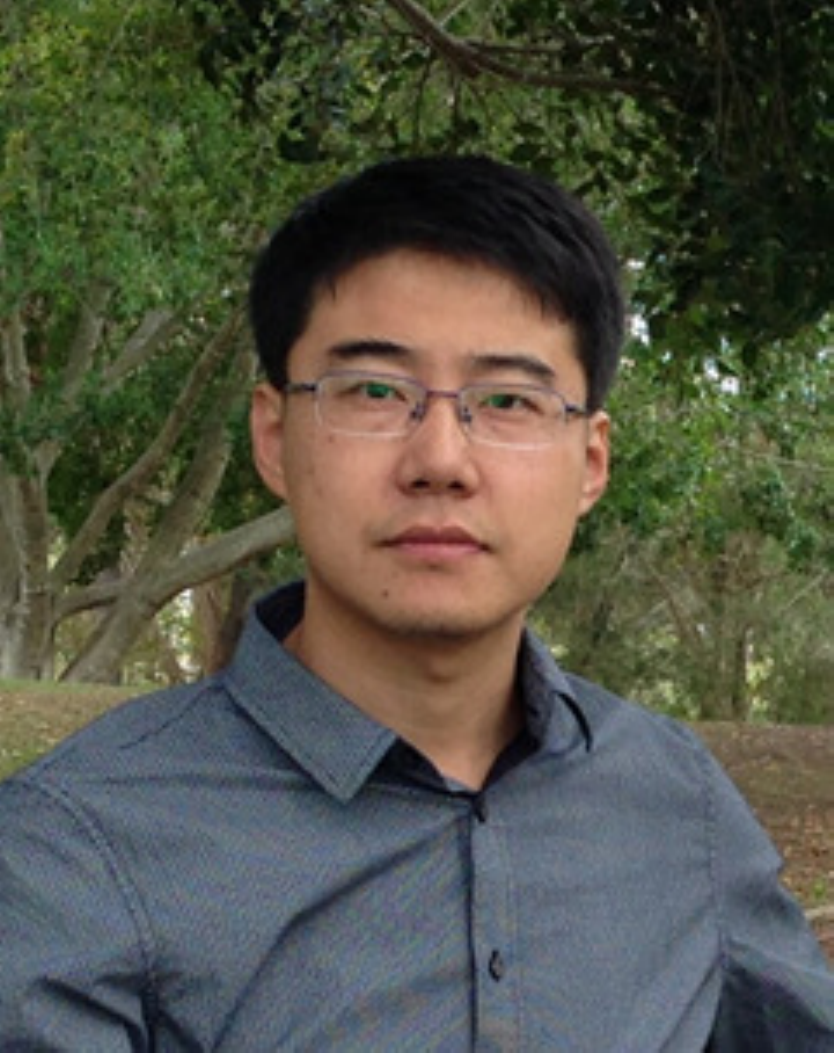}}]{Yang Yang} received the bachelor’s degree from
Jilin University in 2006, the master’s degree from Peking University in 2009, and the Ph.D. degree from The University of Queensland, Australia,
in 2012, under the supervision of Prof. H. T. Shen and Prof. X. Zhou. He was a Research Fellow under the supervision of Prof. T.-S. Chua with the National University of Singapore from 2012 to 2014. He is currently with the University of Electronic Science and Technology of China.
\end{IEEEbiography}

\begin{IEEEbiography}[{\includegraphics[width=1in,height=1.25in,clip,keepaspectratio]{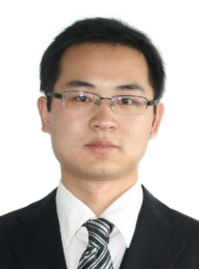}}]{Huimin Lu}
Huimin Lu received double M.S. degrees in Electrical Engineering from Kyushu Institute of Technology and Yangzhou University in 2011. He received a Ph.D. degree in Electrical Engineering from Kyushu Institute of Technology in 2014. From 2013 to 2016, he was a JSPS research fellow (DC2, PD, and FPD) at Kyushu Institute of Technology. Currently, he is an assistant professor in Kyushu Institute of Technology and an Excellent Young Researcher of Ministry of Education, Culture, Sports, Science and Technology-Japan. His research interests include artificial intelligence, computer vision, computational imaging, deep-sea observing, internet of things and robotics.
\end{IEEEbiography}





\vfill


\end{document}